# Credit Card Fraud Detection in the Nigerian Financial Sector: A Comparison of Unsupervised TensorFlow-Based Anomaly Detection Techniques, Autoencoders and PCA Algorithm

Jennifer Onyeama
Onyeama.jennifer@esut.edu.ng
Enugu State University of Science and Technology

## ABSTRACT

Credit card fraud is a major cause of national concern in the Nigerian financial sector, affecting hundreds of transactions per second and impacting international e-commerce negatively. Despite the rapid spread and adoption of online marketing, millions of Nigerians are prevented from transacting in several countries with local credit cards due to bans and policies directed at restricting credit card fraud. Presently, a myriad of technologies exist to detect fraudulent transactions, a few of which are adopted by Nigerian financial institutions to proactively manage the situation. Fraud detection allows institutions to restrict offenders from networks and with a centralized banking identity management system, such as the Bank Verification Number used by the Central Bank of Nigeria, offenders who may have stolen other people's identities can be back-traced and their bank accounts frozen. This paper aims to compare the effectiveness of two fraud detection technologies that are projected to work fully independent of human intervention to possibly predict and detect fraudulent credit card transactions. Autoencoders as an Unsupervised Tensorflow-Based Anomaly Detection Technique generally offers greater performance in dimensionality reduction than the Principal Component Analysis, and this theory was tested out on Nigerian credit card transaction data. Results demonstrate that autoencoders are better suited to analyzing complex and extensive datasets and offer more reliable results with minimal mislabeling than the PCA algorithm.

## 1. INTRODUCTION

In recent years, the Central Bank of Nigeria has continuously enacted a wide range of proactive and highly restrictive policies to curb the rise of credit card fraud. From limiting international transaction amounts to $100 monthly and further reducing the limit to $20 per card, the country has struggled to mitigate the effects of one-click e-payment channels on transactional integrity.

Credit card fraud is a relatively broad terms that describes a situation when unauthorized users gain access to an individual's credit card information in order to make purchases, other transactions, or move funds to another destination [1]. Essentially, an individual commits credit fraud when they use another's identity and creditworthiness to obtain credit or purchase goods and services without the intention of repaying the debt [2]. The unprecedented rise of financial technology platforms, otherwise known as fintech apps, is a leading cause of credit card fraud in the country as there are now more informal access channels with minimal connection to the country's banking verification systems [3]. Another significant enabler of the situation in the country is the use of Point-of-Sale machines, otherwise known as POS machines, where card transactions are performed with hand-held machines in the most unregulated environment areas with requirements for machine acquisition. Over 12.2 billion Naira was lost to fraudulent credit card activity in 2023 [4].

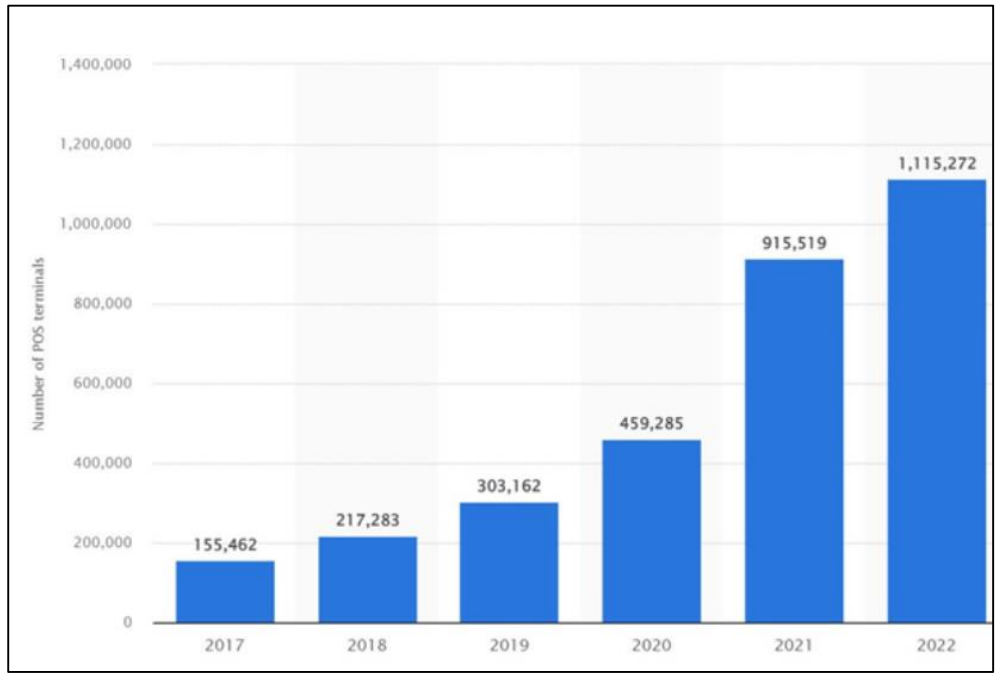

Fig 1: Increasing number of POS terminals in Nigeria over a five-year timeline

Research over the year show that credit card fraud is a global problem creating vacuums in the financial sectors of dozens of countries around the world [5, 6, 7]. Consequently, credit card fraud detection technologies have progressively constituted an essential research area of high interest to many scholars and experts. With the rapid spread and growing relevance of machine learning and big data analysis, more technologies, algorithms and detection methods are uncovered and fine-tuned by researchers across various levels of complexity.

The biggest challenge to research in credit card fraud detection in Nigeria is access to actionable, reliable, and uncompromised datasets for study. This data is most reliable when accessed from traditional financial institutions in the country. However, preliminary assessments show that while policies are enacted and technologies are deployed by this institution is variably efficient attempts to curb credit card fraud, limited efforts have been invested in documenting the instances or at the very least, making the collected data accessible to academic researchers for studies.

Another challenge is the potential for imbalance of data where the number of legitimate transactions is always much larger than the number of fraudulent transactions [8]. This results in poor reflection of the problematic situation upon visualization. With machine learning algorithms, this imbalance also leads to a problem called the "curse of dimensionality", characterized by a situation where the number of records are insufficient for the number of features [9]. Over-fitting and extremely poor generalization are often the results of training with an inadequate or highly imbalanced dataset [10, 11]. Other times, a higher dimensionality to secure a larger number of fraudulent records can cause elongated training periods and ultimate burnout of the system in limited research budgets. Therefore, dimensionality reduction techniques are well-recommended for consolidating stable middle grounds in the iterations and visualizing reliable data.

This paper aims to explore the efficiency of unsupervised machine learning algorithms using Nigerian credit card fraud dataset. It explores autoencoders with code construction from first principles and the Principal Component Analysis (PCA) algorithm with an objective to establish autoencoders as the superior and more reliable fraud detection technique. This paper shows that PCA performs a learning process on a linear transformation which extends or transports the data into another spatial dimension, where vectors of projections are defined by variance of the data [12]. Through a sub-process of restricting the dimensionality to a usually lower number of components that account for most of the variance of the data set, dimensionality reduction is eventually achieved. Autoencoders are neural networks trained using back propagation to reconstruct the input that can be used to reduce the data into a low dimensional latent space by stacking multiple non-linear transformations or layers [13]. They are constructed an encoder-decoder architecture. The encoder maps the input to latent space and decoder reconstructs the input.

The main contributions of this paper are as follows:

- To focus the ongoing research in the credit card fraud detection scene in the cumulative West African financial setting, specifically Nigeria.
- To realign credit card fraud detection as an anomaly detection situation.
- To build an autoencoders system from first principles using Python Programming Language and train it to accurately predict fraudulent transactions. Construct a PCA code base to label fraudulent truncations in a dataset.
- To show how experimental results on a GitHub Dataset with a test set and training show that autoencoders outperforms the PCA algorithm and produces more reliable output for accurate fraud detection.

Programming Language Used:

- Python. Frameworks – TensorFlow, Keras, SciKitLearn

## 2. Literature Review of Related Works

With the rise of global adaptation of credit cards, much so to the apogee of cashless economies thriving solely on the use of credit cards and the accompanying RFID chips, the risk of fraud and malignant truncations are also on the rise [14]. Hundreds of scholars around the world have delved into in-depth research over the years; developing algorithms, technologies, and proposing mechanisms for efficient detection of credit card fraud in national, local, and sub-level financial systems.

Algorithms such as logistic Regression, Genetic Algorithm, Artificial Immune System Model, Naive Bayes, Random Forest, K Nearest Neighbor, Gradient Boosting, Support Vector Machines, neural network algorithm, and the PCA have been explored over the years by many researchers aiming to develop technologies and improve novel algorithms for credit card fraud detection [15, 16, 17]. In 2013, Patel et al proposed a system for credit card fraud detection using Genetic Algorithm (GA) [18]. The genetic algorithm is an optimization technique that mimics the natural evolution processes to select the best characteristics for adaptation a higher likelihood for survival and reproduction. The algorithm used in this study was aimed at obtaining better solutions as time progresses. It operated on the principle of limit dependency, noting that when a card is copied or stolen or lost and captured by fraudsters it is usually used until its available limit is depleted. Thus, rather than the number of correctly classified transactions, a solution which minimizes the total available limit on cards subject to fraud is more prominent. The main objective was to significantly mitigate the occurrence of false alerts using genetic algorithm where a set of interval valued parameters are optimized.

In 2014, Halaviee et al developed a novel model for credit card fraud detection based on the translating the AIS processes into machine learning, simulating natural immune system functionality to protect and prevent anomalies from occurring in systems [19]. This machine learning algorithm worked off the principle of immune system adaptability in humans, where the AIS addresses detecting non-self-cells, imitating the functions of human body which occurs during generating detector cells, detecting non-self-cells, and cleaning the body from non-self while learning its pattern. The algorithm aimed to develop a system where unwanted signals are immediately detected upon entrance or infiltration of the system. The final developed system attempted to improve a previously introduced algorithm in several categories to attain higher precision. It finally proposed a novel implementation model for the method in order to reduce training time.

In 2019, Fiore et al explored the use of generative adversarial networks (GAN) to improvise effectiveness in the classification process of credit card fraud detection [20]. A GAN consists of two feed-forward neural networks, a Generator *G* and a Discriminator *D* working in an adversarial setting with each other, with *G* producing new elements and its opponent *D* instantly evaluating their usefulness and authenticity. This study aimed to generate a large number of reliable and useful examples of the minority class that can be used to re-balance the training sets used by the binary classifier. A careful experimental evaluation showed that a classifier trained on the augmented set largely outperforms the same classifier trained on the original data, especially as far as sensitivity is concerned, resulting in a really effective credit card fraud detection solution. While the framework was developed in the context of credit card fraud detection, the study noted that it could be useful when extended to other application domains.

In 2020, Madhav et al proposed a system based on the PCA and K-Means for the analysis of credit card fraud data [21]. PCA is used for extracting multiple features simultaneously. Firstly, PCA is applied on the fraud dataset. Then PCA creates new features from original features. By making use of these new features, the relation between multiple attributes or associated attributes can be represented in the form of graphs for analysis purpose. The KMeans algorithm is applied on the features which is used to identify the similarity index of associated attributes.

The Random Forest algorithm is one of the most widely researched fraud detection algorithms in the past two decades [22]. It is also the basis of many patented technologies and commercialized products for large-scale detection. In many cases, it is supported by other algorithms for increased functionality and faster runtimes in enterprise-level operations. In 2022, Devi at al Combined RFA technology with the PCA algorithm to develop a web-based technology for inputting a dataset and detecting fraudulent transactions [23]. In 2023, Afiriye et al presented a study which compared three classification and prediction techniques, Decision Tree, Logistic Regression, and Random Forest, conclusively describing the Random Forest Algorithm as the most suitable supervised learning technique for credit card fraud detection [24]. The researchers balanced the dataset prior to generating the models using the under-sampling technique, to ensure that the model does not favor solely the majority class and prevent over fitting the model to the data. With an AUC value of 98.9% and an accuracy value of 96.0%, the Random Forest model performed better than the other two models, making it the most suitable model for predicting fraudulent transactions.

In 2023, Thorat et al experimented the efficiency of the Support Vector Machine (SVM) algorithm in comparison to other machine learning models in fraud detection [25]. Their proposed system with the SVM model of real databases acquired a maximum accuracy of 99.9%. The Artificial Neural Network (ANN) comes with 97.32% accuracy while Hidden Markov Model (HMM) has 94.7% accuracy. ANN has a high processing time and excessive training for large neural networks, difficult to set up and run. Also, Bayesian Networks need excessive training and have 96.52% accuracy.

## 3. Methodology

In this paper, fraud detection in Nigerian credit card transaction data is treated as an anomaly detection problem. The major objectives of the methods applied is to determine the most reliable and well-fitted unsupervised detection technique. It is a layered comparison of Autoencoders built from scratch and the Principal Component Analysis algorithm. The visualization formats of both techniques are vastly different. However, the precision of data represented contributes to the advantages of one method over the other, as is discussed in the following sections.

### 3.1 Autoencoders

Autoencoders are a type of neural network architecture designed and primarily used for unsupervised learning and data compression [26]. They are categorized as a type of unsupervised learning technique that using back propagation to learn the compressed information in raw data. The main objective of autoencoders is to perform dimensionality reduction on a dataset, a technique used to reduce the number of features in a dataset while retaining as much of the important information as possible, preventing over-fitting and elongated training periods [27].

Autoencoders yield a recreation of the input. The autoencoders comprises of two sub-systems: an encoder and a decoder. During training, the encoder learns a set of training, a process identified as a latent representation, from the data it receives. At the same time, the decoder undergoes a similar training process to reconstruct the input. The autoencoders can at that point be connected to perform predictive operations on future input data. Autoencoders are exceptionally generalizable and can be utilized on distinctive information sorts, including picture data, time series, and text.

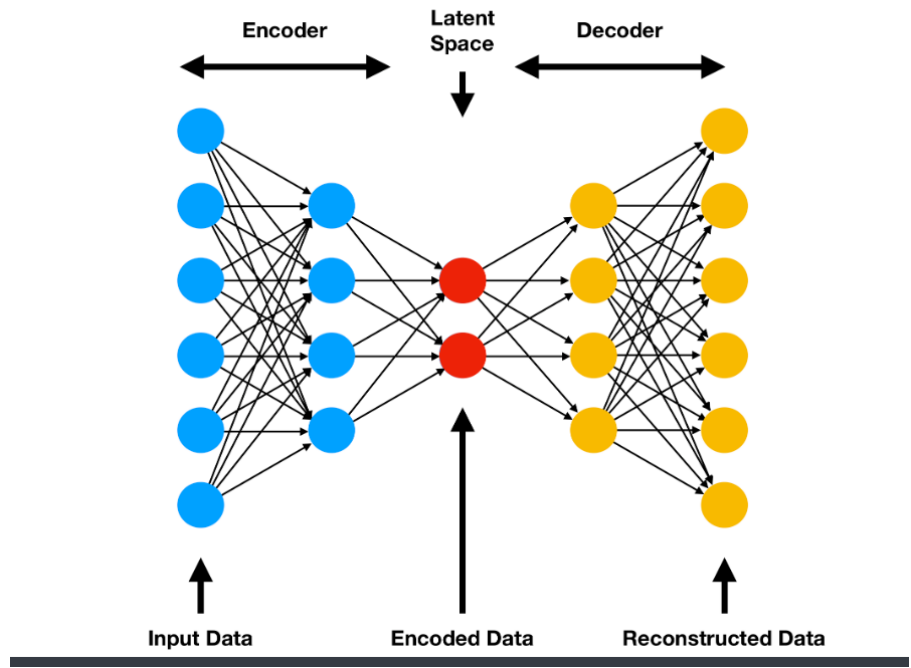

Fig 2: How Autoeconders work

The principle of operation of autoencoders is similar to principal component analysis (PCA). The major difference lies in the non-linearity of the activation function used in autoecnoders. If the AF was linear within each processing layer of the network, the latent variable of the autoencodcers would be exactly the same as the major functions in the PCA algorithm. Generally, the activation function used in autoencoders is non-linear, typical activation functions are ReLU (Rectified Linear Unit) and sigmoid.

Autoencoders can be segmented into three parts that can be represented mathematically as:

$$\emptyset : x \rightarrow f$$
$$\psi : f \rightarrow x$$
$$\emptyset, \psi = \arg \min \| x - (\psi \circ \emptyset) X \|^2$$

**Encoding**: The encoder function, denoted by $\phi$, connects the input data X, to latent space F. The decoder is represented by $\psi$ which maps the latent space F to the output. In the autoencoders architecture, the output is the same as the input, denoting an attempt to recreate the original input following a process of generalized non-linear compression.

The encoding network can be represented by the standard neural network function passed through an activation function, where **z** is the latent dimension;

$$Z = \sigma(Wx + b)$$

**Decoding**: The decoding network is represented similarly but with different a weight, bias, and potentially different activation functions.

$$X' = \sigma'(W'z + b')$$

The loss function is expressed in terms of the functions above, and it is this loss function that is utilized in training the neural network through the standard back propagation procedure.

$$L(X,X') = \|x-x'\|^2 = \|x - \sigma'(W'(\sigma(Wx + b)) + b')\|^2$$

Another common loss function used for autoencoders is the mean squared error (MSE) loss, which is calculated as:

$$L = 1/N * sum((x - x')^2)$$

A reduction of the loss function through the gradient descent enables the autoencoders system to learn to encode and decode the input data efficiently. The compressed representation is then used in anomaly detection and in our case, credit card fraud detection. One major objective of the autoencoders system is to select the encoder and decoder functions in such a way that minimal information is required to encode the dataset such that it be can regenerated on the other side.

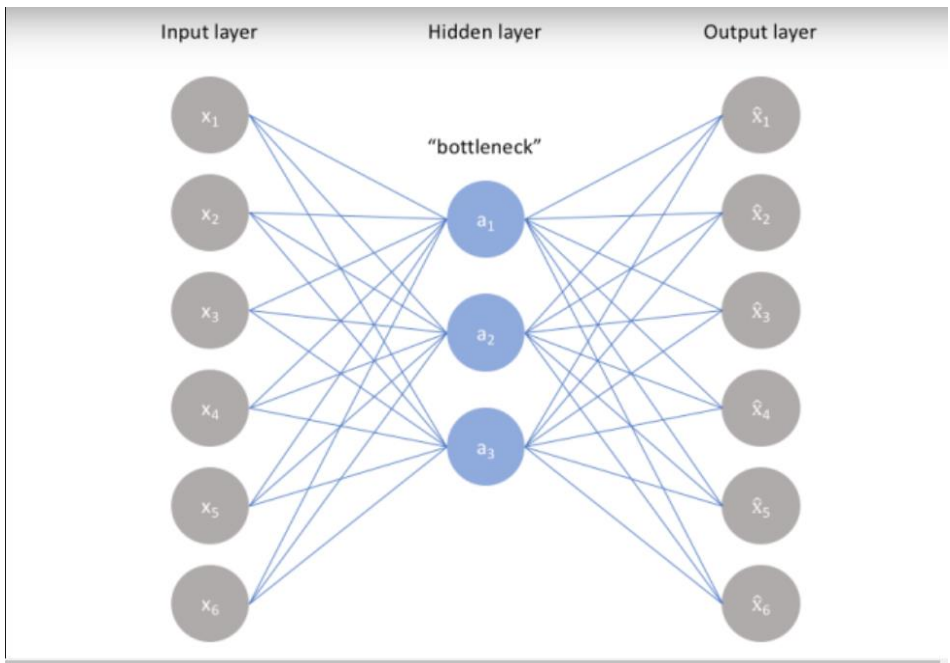

Fig 3: Autoencoders bottleneck depiction

### 3.1.1 Autoencoders Code Construction

For the purpose of this paper, the Python programming language was used to build the autoencoders. The dataset was retrieved from a GitHub repository where Nigerian credit card fraud data retrieved from financial institution was stored by a research team. Using the Keras API built on top of TensorFlow for model building, imported into Jupyter Notebook as shown below:

> *from tensorflow.keras.models import Model*
>
> *from sklearn.model_selection import train_test_split*
>
> *from sklearn.preprocessing import MinMaxScaler, StandardScaler*

The train and test files are combined using the !cat function and saved in another file while the dataset is loaded onto the Notebook. A header is assigned to the dataset to prevent it from automatically mapping the first row of values as its header. Prior to classifying the data as normal and anomaly (fraud), the data is classified into the train and test sets. A normalization process is carried out to ensure that a common scale is available to the neural network for its processing.

> *x_train, x_test, y_train, y_test = train_test_split(df.values, df.values[:,0:1], test_size=0.2, random_state=111)*
>
> *scaler = MinMaxScaler()*
>
> *data_scaled = scaler.fit(x_train)*
>
> *train_data_scaled = data_scaled.transform(x_train)*
>
> *test_data_scaled = data_scaled.transform(x_test)*

The test data and training data are split into normal and anomaly sets. Now the training dataset has its own normal and anomaly data frame. This is an 80-20 split. The training is carried out only with the normal dataset while the anomaly is used as a validating reference.The first three columns of the dataset are the plotted. Notice the cascading of the graphs and moving into normalcy domain.

### 3.1.2 Modelling the AutoEncoder.

This paper uses the sub-classing or model sub-classing method to model the AutoEncoder, against the sequential modelling method provided by Keras API on top of TensorFlow. Subclassing is used because it separates the operations of the encoder and decoder in an easier process. To instantiate, subclassing offers the ability to use only the encoder for compression if it is needed, exclusive of the decoder. A class and constructor were created in the same way of decreasing layer order and 8 units are bottleneck layers. Where an encoder performs a down-sampling of the data, the decoder performs an up-sampling.

An early stopping phenomenon is used to terminate the training process in an unexpected situation where the decrease of the loss function is not stopped after iterations. The mean absolute error is used as the loss function in this class. The model was compiled using the adam optimization method and the train data passed through it in two iterations. The matplotlib library in Python was used to visualize the graphs and data at every stage in the study.

## 3.2 Principal Component Analysis

Principal component analysis (PCA) is dimensionality reduction technique used on high-volume datasets. The operating idea is to extract the principal components that hold the most relevant pieces of data and delete the least important once, thereby performing a de-noising operation on a dataset while preserving the most useful information. It is used on very enormous collections of data. The outcome of the PCA algorithm is a significant and massive reduction of a larger dataset into a much smaller and compact format for processing. Consequently, the compression of this data causes a lot of the precision to be lost. PCA aims to improve the efficiency of data analysis and visualization which is generally simpler with smaller datasets.

Running the PCA algorithm produces new feature identified as principal components and there must be an accurate correspondence to the linear order of the initial features. Orthogonal characterization limits similarity between any pair of variables. From 1 to n, the significance of each component diminishes. This indicates that the number one PC is the most important, while the number "n" PC is the least significant. PCA is a highly adaptable approach to analyzing datasets that may include multicollinearity, absent numbers, categorized data, and imprecise metrics, among others. The objective is to extract essential details from the data and convey it as an array of overarching indices known as principal components.

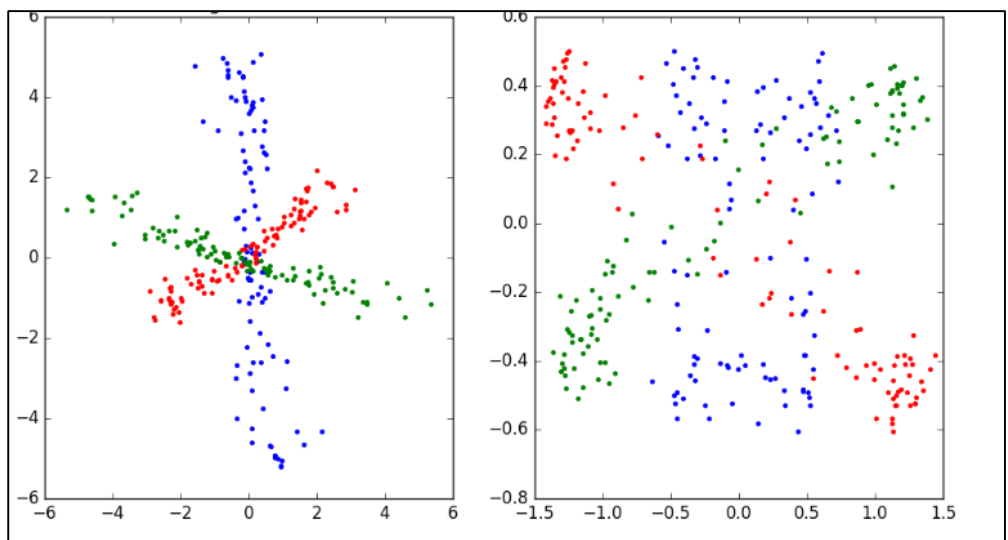

Fig 4: Graphical representations of PCA

Mathematically, the steps involved in PCA were represented as:

**Standardization**: This involves a calculation of the mean value of the dataset dimensions excluding the labels. The data is scaled to ensure an equal contribution of variables. The formula below is a simple mathematical expression where z is the scaled value, x is the initial, and *mu* and *sigma* are mean and standard deviation, respectively.

$$Z = (X - \mu) / \sigma$$

**Covariance Matrix Computation**: Covariance of X and Y is calculate as thus:

$$cov(X,Y) = \frac{1}{n-1}\sum_{i=1}^{n} (X_i - \bar{x})(Y_i - \bar{y})$$

The above formula is used to find the covariance matrix. Also, the result would be a *square matrix of x*x dimensions.* Covariance matrix in python is represented as follows:

*cov_matrix = np.cov(df_num_scaled.T)*

*print('Covariance Matrix n%s', cov_matrix)*

**Computation of Eigenvectors and corresponding Eigenvalues**: An eigenvector is a non-zero vector that changers by a scalar value. An eigenvalue is then produced when this conversion occurs. If a matrix M, exists, is eigenvector is represented as follows:

$$M\bar{v} \rightarrow \lambda\bar{v}$$

**Selecting the eigenvectors with the highest eigenvalues**: This eigenvector sort is performed in decreasing order of values with the total number representing the number of dimensions required in the new dataset to be produced.

### 3.2.1 PCA Code Construction

Using PCA, unsupervised learning technique was used to identify fraudulent transactions in the dataset. The dataset was loaded into the notebook and the samples separated by calls to represent legitimate and illegitimate transactions. The next step in the process involved delisting the time and class columns from the code segment.

```
import pandas as pd

df = pd.read_csv('Data/dataset.csv')

df.head()

# Separate the samples by class

legit = df[df['Class'] == 0]

fraud = df[df['Class'] == 1]

# Drop the "Time" and "Class" columns

legit = legit.drop(['Time', 'Class'], axis=1)
```

fraud = fraud.drop(['Time', 'Class'], axis=1)

Following this step is using the algorithm to drastically reduce the number of dimensions in the dataset, fitting into legitimate transactions only. Problematically, bits of information dropped out of the dimensions during the transformation. Finally, the loss for each row in the legitimate and fraudulent transaction datasets.

>    Losses for legitimate transactions < 200

>    Losses for fraudulent transactions > 200

## 4. Results and Conclusions

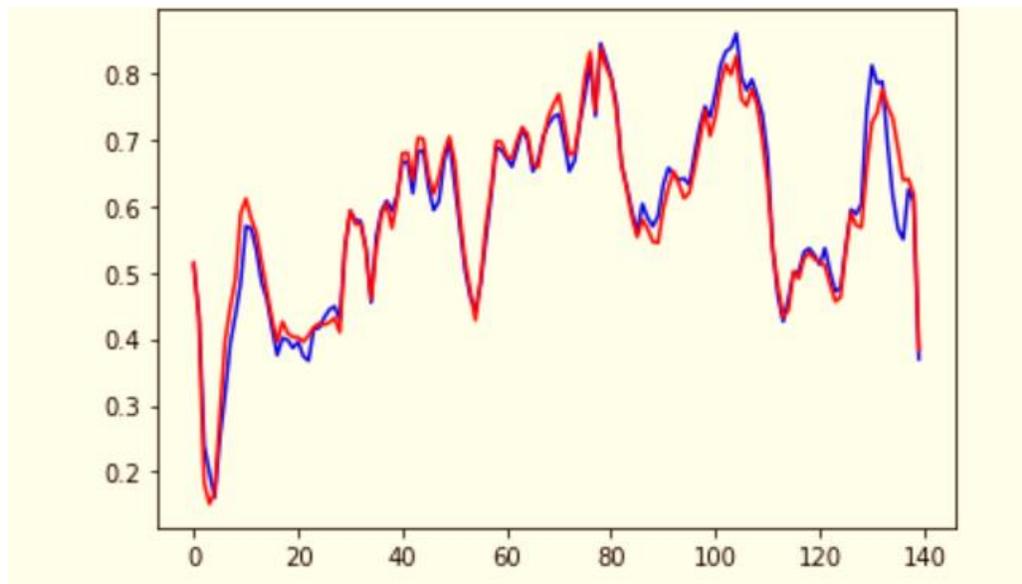

Fig 5: Autoencoders Model performance on Normal data

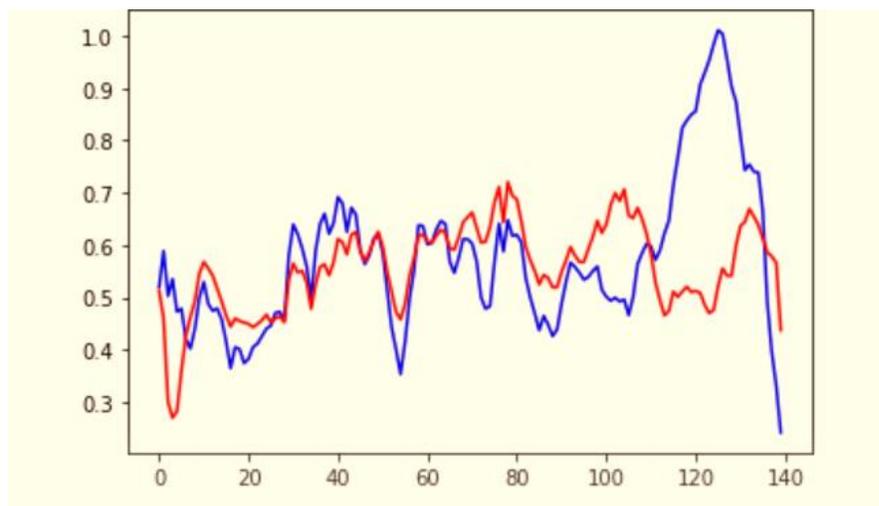

Fig 6: Autoencoders performance on fraudulent transactions

## 4.1 Results of the Autoencoders Iteration Processes:

To identify all values embedded in the anomaly data, the follower TensorFlow function is used:

    preds = tf.math.less(train_loss, threshold)

    tf.math.count_nonzero(preds)

Out of 2454 total records, the model accurately predicts 2346 values correctly. It is estimated to be 95% accurate.

A TensforFlow "greater" function is applied to find the count of values that are greater than a threshold which is anomalies.

    preds_a = tf.math.greater(train_loss_a, threshold)

    tf.math.count_nonzero(preds_a)

From 2135 total records, the model precisely predicted 2087. The final model will estimable have a 0-95% accuracy in predicting the new points.

## 4.2 Results of the PCA iteration processes:

The findings show that the PCA results did not perform as optimally as the Autoencoders. Despite the disparity, the model captured 50% of the fraudulent transactions while mislabeling just 3 out of 2454 legitimate transactions.

Estimably, this is an error rate of less than 0.12% for legitimate transactions, compared to 0.007% for the supervised-learning model. Further iterations revealed that the accuracy can be improved by experimenting with different values. A lower threshold will improve the model's capability at capturing illegitimate transactions. However, it will most likely mislabel more legitimate transactions.

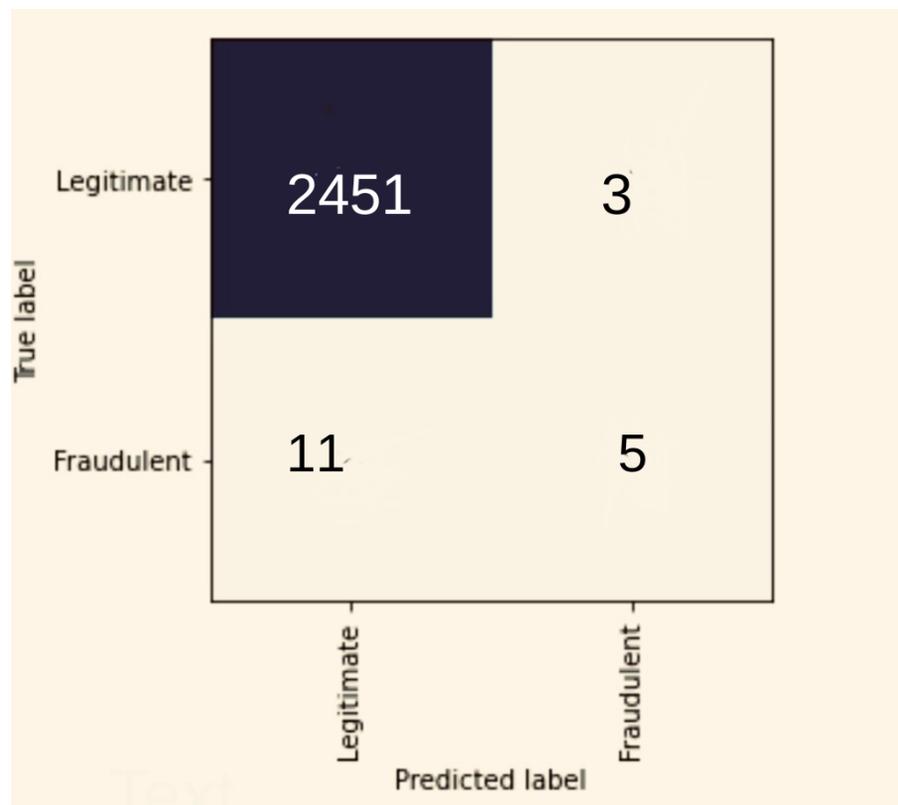

Fig 7: Overall PCA algorithm performance

## 5.0 Final Observations

The final observation of the study revealed that the autoencoders presents a more optimal and more dependable anomaly detection performance that the PCA. The non-linearity of the range of autoencoders makes the system a more reliable and flexible option for fraud detection. PCA is a deterministic and linear method, while autoencoders are stochastic and nonlinear. The Principal Component Analysis always gives the same result for the same data, capturing only linearly structured relationships.

Autoencoders can capture the full range of nonlinear data, producing different results based on the random initialization and optimization processes. With a constant set of parameters that are not subject to sudden or random change, PCA depends on data dimensionality for its analytical processes. Autoencoders performs independent of the data and relies more on the architecture and complexity of the network. The global processing power of PCA allows it to be more computationally efficient. However, Autoencoders consider data paints locally during its processes and thus, it handles complex data more efficiently.

## References

1. Shah, A.; Makwana Y.; Credit Card Fraud Detection. *Research Gate*. 2023. [https://www.researchgate.net/publication/369857378_Credit_Card_Fraud_Detection]